\documentclass[11pt,a4paper]{article}
\usepackage[hyperref]{eacl2021}
\usepackage{times}
\usepackage{latexsym}

\usepackage{microtype}

\aclfinalcopy 

\usepackage{graphicx}
\usepackage{enumitem}

\title{The Great Misalignment Problem in Human Evaluation of NLP Methods}

\author{Mika Hämäläinen \\
  University of Helsinki and Rootroo Ltd \\
  \texttt{mika.hamalainen@helsinki.fi} \\\And
  Khalid Alnajjar \\
  University of Helsinki and Rootroo Ltd \\
  \texttt{khalid.alnajjar@helsinki.fi} \\}

\date{}

\begin{document}
\maketitle
\begin{abstract}
We outline the Great Misalignment Problem in natural language processing research, this means simply that the problem definition is not in line with the method proposed and the human evaluation is not in line with the definition nor the method. We study this misalignment problem by surveying 10 randomly sampled papers published in ACL 2020 that report results with human evaluation. Our results show that only one paper was fully in line in terms of problem definition, method and evaluation. Only two papers presented a human evaluation that was in line with what was modeled in the method. These results highlight that the Great Misalignment Problem is a major one and it affects the validity and reproducibility of results obtained by a human evaluation.
\end{abstract}

\section{Introduction}

There has been a lot of academic discussion recently about different evaluation methods used and their validity \cite{novikova2017we,reiter2018structured,howcroft-etal-2020-twenty,van-der-lee-etal-2019-best}. Reproducibility is an important problem in our field of science and it is not currently archived in human evaluation, as some researches have found that trying to reproduce a human evaluation gives different results \cite{hamalainen2020automatic,mieskes2019nlp}. 

However important reproducibility is, we have identified an even more severe problem in human evaluation. We call this problem \textit{the Great Misalignment Problem} that is a mismatch between a problem statement, a proposed model and a proposed evaluation method.

It is typical in the field of NLP to work with ill-defined problems. For instance, many machine translation papers \cite{roest-etal-2020-machine,chen-etal-2020-content,talman-etal-2019-university} do not extensively define what they mean by translation, a topic that has multiple definitions in translation studies \cite{hermans2014manipulation,reiss1989text,lederer2014translation}, but merely take it for granted and focus on proposing systems that achieve high scores in an automatic evaluation metric such as BLEU \cite{papineni2002bleu}.

For as long as you work with a problem the solution of which you can objectively measure by automated metrics, the role of a problem definition is not that important. The situation changes, however, when your main evaluation method is a subjective human evaluation. The reason for this is simple: only when you have defined the problem clearly, can you derive the questions and methods for a human evaluation (c.f. \citealt{alnajjar2018master,Jordanous2012}). When one does not have a clear understanding of the problem one seeks to solve, the evaluation is usually not representative of the problem, thus they are misaligned.

The Great Misalignment Problem is not just about the misalignment between the problem definition and the evaluation, but also the proposed solution, let it be rule-based, algorithmic or a machine learning model. We can often see that the solution itself has very little to do with the human evaluation methods used.

In this paper, we study the Great Misalignment Problem (alignment of a problem definition, method and human evaluation) by surveying papers published in ACL 2020 that use human evaluation. We focus on ACL since it is supposed to be the most prestigious conference in the field. For courtesy reasons, we anonymize the papers surveyed, except Paper 3~\cite{mohankumar-etal-2020-towards} which was the only paper that did not exhibit the Great Misalignment Problem. We do not want single anyone out with our critique as that is not the goal of our paper.

\begin{table*}[!ht]
\centering
\begin{tabular}{|l|l|l|l|l|l|}
\hline
         & Definition  & \begin{tabular}[c]{@{}l@{}}Method in line\\ with the definition\end{tabular} & \begin{tabular}[c]{@{}l@{}}Evaluation in line\\ with the definition\end{tabular} & \begin{tabular}[c]{@{}l@{}}Evaluation in line\\ with the method\end{tabular} & \begin{tabular}[c]{@{}l@{}}Evaluation in line\\ with the topic\end{tabular} \\ \hline
Paper 1  & Theoretical & No                                                                       & No                                                                           & No                                                                           & No                                                                          \\ \hline
Paper 2  & Absent      & No                                                                       & No                                                                           & \textbf{Yes}                                                                          & \textbf{Yes}                                                                         \\ \hline
Paper 3  & ML          & \textbf{Yes}                                                                      & \textbf{Yes}                                                                          & \textbf{Yes}                                                                          & \textbf{Yes}                                                                         \\ \hline
Paper 4  & Absent      & No                                                                       & No                                                                           & No                                                                           & No                                                                          \\ \hline
Paper 5  & Absent      & No                                                                       & No                                                                           & No                                                                           & \textbf{Yes}                                                                         \\ \hline
Paper 6  & Absent      & No                                                                       & No                                                                           & No                                                                           & No                                                                          \\ \hline
Paper 7  & Math        & \textbf{Yes}                                                                      & No                                                                           & No                                                                           & No                                                                          \\ \hline
Paper 8  & Theoretical & \textbf{Yes}                                                                      & No                                                                           & No                                                                           & \textbf{Yes}                                                                         \\ \hline
Paper 9  & Absent      & No                                                                       & No                                                                           & No                                                                           & No                                                                          \\ \hline
Paper 10 & Absent      & No                                                                       & No                                                                           & No                                                                           & No                                                                          \\ \hline
\end{tabular}
\caption{The Great Misalignment Problem in the papers surveyed.}
\label{tab:survey-results}
\end{table*}

\section{Surveying the Great Misalignment Problem}

We filter all papers that have the words ``human'' and ``evaluat*'' or ``judge*'' in their abstract. This way, we can find papers mentioning human evaluation, human evaluators and so on. We include all papers published in the ACL 2020\footnote{https://www.aclweb.org/anthology/events/acl-2020/} (excluding workshops) in the search. We sort these papers (79 in total) at random and take the first 10 papers that actually have used human evaluation, as some of the papers mentioned human and evaluation, but did not conduct a human evaluation. We did not consider papers that suggested automated evaluation metrics based on correlation with human evaluation as their main contribution. Human evaluation is most common in natural language generation as 8 out of 10 papers deal with NLG.

The papers, we considered for evaluation in terms of human evaluation, presented automatic evaluation metrics in addition to human evaluation. For all the 10 papers, we looked at the following questions:

\begin{itemize}[noitemsep,topsep=0pt]
  
  \item How is the problem defined and narrowed down?
  \item Is the proposed method in line with the definition?
  \item Is the evaluation in line with the definition?
  \item Is the evaluation in line with what was modeled by the method?
  \item Is the evaluation in line with the overall topic of the paper?
\end{itemize}

As an example, if a paper proposes a model for poem generation and does not define what is meant by poem generation, we consider the definition to be absent. A simple statement of the topic is not enough as there are nuances to poem generation such as rhyme, meter, metaphors, symbolism, personification and so on. If a paper presents a more narrowed-down definition and this definition is followed in the method proposed, we consider the two to be in line.

Evaluation is in line with the definition, if the evaluation questions reflect the different aspects that were defined important in the problem definition. For the evaluation to be in line with the model, it should evaluate what the model was designed to do. If for example, a poem generator model takes meter and rhyme into account, but it is evaluated based on fluency and poeticness, the method and the evaluation are not in line. For them to be in line, meter and rhyme should have been evaluated. The evaluation can be in line with the overall topic of the paper: for example, evaluating poeticness is in line with poem generation.

The results of our survey can be seen in Table \ref{tab:survey-results}. As we can see, almost all papers had the Great Misalignment Problem except for one paper, Paper 3. Unlike the rest of the papers surveyed, this particular paper did not try to solve an NLP problem per se, but rather focused on studying the attention models used in LSTM neural networks. Therefore its problem definition, method and human evaluation focused on the attention models rather than any NLP problems.

Paper 7 presented a very explicit mathematical statement for the problem they were to solve in the paper. Although, this is very specific to the implementation the authors had, it is still better than a completely absent definition as seen in the majority of papers that took an abstract level topic for granted and solved it with some method.

Paper 1 and Paper 8 used an existing theory to narrow down the topic. Paper 8 did this in a good way, as their implementation followed exactly the notions defined by the theory they used. However, Paper 1 merely mentioned a theory for their definition, completely ignoring it in the implementation of the method and in the evaluation.

Papers 2, 4-6 and 9-10 do not provide any definition for the problem they are trying to solve, but rather take the definition for granted. Therefore their evaluation cannot be in line with the definition either, as no definition was provided, but in some cases the evaluation was at least in line with the overall topic of the paper, although this was not always the case.

Only Paper 3 had their evaluation in line with the definition and only Paper 3 and Paper 2 had their evaluation in line with what was modeled in the method. This is very concerning, as it highlights how little the evaluation questions used had to do with what was actually done in the papers. On a more positive note, Papers 2, 3, 5 and 8 at least have their evaluation in line with the topic of the paper, however this means that are 6 papers the evaluation of which is not in line with the topic.

Table \ref{tab:sample-size} show how many samples (different outputs by a system) were evaluated. We can see that there is a lot of variety in this respect in the papers surveyed, but half of the papers have evaluated from 100 to 500 samples. The situation gets even more complicated when we look at the results reported in Table \ref{tab:judges-per-sample}. Here, we can see that there is a lot of variety in how many human evaluators evaluated each sample. Two of the papers did not report this at all.

\begin{table}[]
\small
\centering
\begin{tabular}{|l|c|}
\hline
\textbf{\# of Samples} & \textbf{N} \\ \hline
Less than 20           & 1          \\ \hline
100-500                & 5          \\ \hline
500-1000               & 2          \\ \hline
1000-1500              & 2          \\ \hline
\end{tabular}%
\caption{Number of samples produced by the method that were evaluated.}
\label{tab:sample-size}
\end{table}

\begin{table}[]
\small
\centering
\begin{tabular}{|l|c|}
\hline
\textbf{Judges per sample} & \textbf{N} \\ \hline
Not Given                 & 2          \\ \hline
1-3                       & 2          \\ \hline
4-5                       & 3          \\ \hline
6-10                      & 2          \\ \hline
Above 10                  & 1          \\ \hline
\end{tabular}
\caption{Number of human evaluators used per sample produced by the method.}
\label{tab:judges-per-sample}
\end{table}

\section{Discussion}

A direct implication of the Great Misalignment Problem is that the results of any human evaluation cannot be reproducible as they are measuring something else than what was modeled in the proposed solution. Therefore, any results obtained by the human evaluation can only be due to some other variable such as the data used in training, a bias in the often too small evaluation sample or a bias in the often too few evaluators.

Furthermore, many factors affect the quality of the human evaluation. For instance, forcing the evaluators to provide answers to questions that they do not know how to answer without giving them the possibility to skip such questions could introduce noise in the evaluation data. On the contrary, some unfaithful evaluators (scammers) might abuse such an opportunity to finish the survey effortlessly and in a short time by submitting valid answers, i.e. ``I do not know''. 

Some surveying platforms support defining criteria to discard scammers, such as test questions or a minimum response time. Test questions are greatly useful to enhance the quality of answers. However, when used for evaluating subjective tasks they would add a bias as evaluators must share the same opinions of the authors or, else, they will be rejected from continuing the survey. The minimum response time is there to eliminate scammers who answer promptly without even reading the questions.

Other similar criteria exist, e.g., language and geographical restrictions that might aid in finding competent evaluators, the ordering of samples when presented side by side, and the bias of providing a single answer consistently to different questions~\cite{90fcde3a2b7e4813b3f74d3fb9b6b371}. This just to show that many factors regarding the human evaluation setup contribute massively to the quality of the evaluation. There is no one fixed or correct way to conduct all human evaluations, but researchers in the field should consider such biases and aim towards reducing them in addition to revealing the full details of the evaluation setup and the intuition behind it to the reader to allow reproducibility of the scientific work. Unfortunately, none of the papers surveyed described the human evaluation conducted in a clear fashion, where different biases or threats to the validity of the results would have been made clear.

Our field is very often focused on gaining the state of the art performance from our models. However, when the human evaluation metrics used have little to nothing to do with the problem or the method, knowing what truly is the state of the art becomes less clear. Each system, regardless of their final evaluation score, will have a lot of advantages and disadvantages that do not become evident if the problem they are used to solve is ill-defined. This leads to the problem that evaluation scores are the only way of showcasing the superiority of your system, no matter how unrelated the evaluation scores were to the problem or to your method.

The problem that comes from not evaluating what you have modeled in your method is that you cannot say whether what you modeled actually works as intended. This is especially problematic in the case of NLG, which represents a majority of papers surveyed. Nowadays generating good sounding text is no longer an issue as very generic models such as GPT-2 \cite{radford2019language} can be used to generate many different kinds of text. This leads to the problem that if no clear definition is provided, any method that spits out text will satisfy the requirements, and if the evaluation does not capture anything about how the method was implemented, then it is impossible to tell whether your system actually improved anything but the very surface of the text.

In our own experiments \cite{hamalainen-alnajjar-2019-generating} with human evaluation, we have found that questions that do not measure what has been modeled make it very difficult to say what should be improved in the system and how, although such an evaluation makes the end results look impressive. As \citet{gervas-2017-template} puts it, any feature not modeled in a generative system that happens to be in the output can hardly be a merit of the system, but is in the result due to mere serendipity. To complicate the things, \citet{Veale+2016+73+92} points out that people are willing to read more content into the output of a system than what the system had planned.

To solve these problems, we decided to follow an approach where we defined exactly what we need our system to be able to produce in its output (humorous headlines). In our first paper \cite{alnajjar2018master}, we believed we had solved the problem, only to realize in our follow-up paper \cite{d76230f1c2ad4e9f87cd5f3840ae2742} that the human evaluation results contradicted our own impression of the output produced by the different systems. As it turns out, the evaluation questions were too abstract and left enough room for people to read more into the output.

While our latest trial in solving the issue has been using concrete evaluation questions \cite{hamalainen2019let} that measure exactly what the system was designed to do in order to reduce subjectivity, such an evaluation practice cannot be embraced if there is no alignment between the definition, solution and evaluation. No matter how concrete the evaluation questions are or how sound the evaluation method is in terms of forming a good quantitative questionnaire, an evaluation that neither evaluates the method nor the problem can hardly be meaningful. 

All in all, we have had good experiences when conducting human evaluation in person by printing out questionnaires and presenting them to people. It is not at all difficult to find test subjects who are willing to participate. This way, one can avoid the problem of paid online questionnaires where the motives and skills of the human evaluators is difficult to assess. Furthermore, conducting evaluation this way, opens the evaluation up for criticism and it is easy to get direct feedback from the participants on the test design and its difficulty.

\section{Conclusions}

In this paper, we have described a fundamental issue in human evaluation in the field of NLP. Our initial survey results show that the issue can be found extensively in the papers published in our field. The Great Misalignment Problem makes it impossible to critically assess the advancements in the field, as usually problems papers are trying to solve, are not defined well enough to be thoroughly evaluated by human judges. In addition, if the method proposed does not align well with the problem nor the evaluation, any human evaluation results can hardly be a merit of the method. 

There are several uncontrolled variables involved and based on our survey results, human evaluation is not conducted in the same rigorous fashion as in other fields dealing with human questionnaires such as in social sciences (c.f. \citealt{babbie2015basics}) or fields dealing with evaluation of computer systems such as design science (c.f. \citealt{hevner2004design}). There is a long way for our field to go from here in order to establish more sound and reproducible human evaluation practices.

Narrowing the problem definition down from an abstract definition such as ``poem generation'' or ``diverse dialog generation'' not only helps in understanding the problem from the point of view evaluation, but also makes it possible to ask more meaningful questions while proposing a solution. Such an ideology can be useful also in domains where evaluation is conducted automatically in order to critically assess the validity of the approach and the evaluation method used.

The results presented in this paper are based on only 10 papers published in ACL 2020. The sample seems representative to the general feel of the state of human evaluation in the field, but it is important in the future to survey a larger sample of papers to better understand the problem. While conducting our survey, we also paid attention to other issues in human evaluation such as the fact that the evaluation methods are not usually adequately described in terms of presentation of the evaluation questions (many papers did not report the questions at all), selection of human judges, task instructions and so on. There were huge differences also in the number of human judges from only 3 to 30 judges, and also in the number of samples evaluated.

Our field does not have an established methodology for human evaluation, but at the current stage, the validity of many human evaluation methods is questionable. This is problematic as our field clearly has problems that rely on human evaluation. We do not believe that removing human evaluation altogether in favor of objective evaluation methods is the optimal solution either, as automatic evaluation metrics come with their own problems and biases. In order to reach to better human evaluation practices, a study of human evaluation itself is needed. From our experiences with human evaluation, we can say that it is certainly not a straight forward problem due to a variety of different reasons, the largest of them being subjective interpretation and limited understanding the human evaluators have of the evaluation task, questions and the actual output that is to be evaluated.

\bibliography{anthology,eacl2021}

\begin{thebibliography}{25}
\expandafter\ifx\csname natexlab\endcsname\relax\def\natexlab#1{#1}\fi

\bibitem[{Alnajjar and H{\"a}m{\"a}l{\"a}inen(2018)}]{alnajjar2018master}
Khalid Alnajjar and Mika H{\"a}m{\"a}l{\"a}inen. 2018.
\newblock A master-apprentice approach to automatic creation of culturally
  satirical movie titles.
\newblock In \emph{Proceedings of the 11th International Conference on Natural
  Language Generation}, pages 274--283.

\bibitem[{Babbie(2015)}]{babbie2015basics}
Earl~R Babbie. 2015.
\newblock \emph{The basics of social research}.
\newblock Nelson Education.

\bibitem[{Chen et~al.(2020)Chen, Wang, Utiyama, and
  Sumita}]{chen-etal-2020-content}
Kehai Chen, Rui Wang, Masao Utiyama, and Eiichiro Sumita. 2020.
\newblock \href {https://doi.org/10.18653/v1/2020.acl-main.34} {Content word
  aware neural machine translation}.
\newblock In \emph{Proceedings of the 58th Annual Meeting of the Association
  for Computational Linguistics}, pages 358--364, Online. Association for
  Computational Linguistics.

\bibitem[{Gerv{\'a}s(2017)}]{gervas-2017-template}
Pablo Gerv{\'a}s. 2017.
\newblock \href {https://doi.org/10.18653/v1/W17-3903} {Template-free
  construction of rhyming poems with thematic cohesion}.
\newblock In \emph{Proceedings of the Workshop on Computational Creativity in
  Natural Language Generation ({CC}-{NLG} 2017)}, pages 21--28, Santiago de
  Compostela, Spain. Association for Computational Linguistics.

\bibitem[{H{\"a}m{\"a}l{\"a}inen and
  Alnajjar(2019{\natexlab{a}})}]{hamalainen-alnajjar-2019-generating}
Mika H{\"a}m{\"a}l{\"a}inen and Khalid Alnajjar. 2019{\natexlab{a}}.
\newblock \href {https://doi.org/10.18653/v1/D19-1617} {Generating modern
  poetry automatically in {F}innish}.
\newblock In \emph{Proceedings of the 2019 Conference on Empirical Methods in
  Natural Language Processing and the 9th International Joint Conference on
  Natural Language Processing (EMNLP-IJCNLP)}, pages 5999--6004, Hong Kong,
  China. Association for Computational Linguistics.

\bibitem[{H{\"a}m{\"a}l{\"a}inen and
  Alnajjar(2019{\natexlab{b}})}]{hamalainen2019let}
Mika H{\"a}m{\"a}l{\"a}inen and Khalid Alnajjar. 2019{\natexlab{b}}.
\newblock Let’s face it. finnish poetry generation with aesthetics and
  framing.
\newblock In \emph{Proceedings of the 12th International Conference on Natural
  Language Generation}, pages 290--300.

\bibitem[{H{\"a}m{\"a}l{\"a}inen and
  Alnajjar(2019{\natexlab{c}})}]{d76230f1c2ad4e9f87cd5f3840ae2742}
Mika H{\"a}m{\"a}l{\"a}inen and Khalid Alnajjar. 2019{\natexlab{c}}.
\newblock \href {http://computationalcreativity.net/iccc2019/} {Modelling the
  socialization of creative agents in a master-apprentice setting: The case of
  movie title puns}.
\newblock In \emph{Proceedings of the 10th International Conference on
  Computational Creativity}, pages 266--273, Portugal. Association for
  Computational Creativity.
\newblock International Conference on Computational Creativity ; Conference
  date: 17-06-2019 Through 21-06-2019.

\bibitem[{H{\"a}m{\"a}l{\"a}inen et~al.(2020)H{\"a}m{\"a}l{\"a}inen, Partanen,
  Alnajjar, Rueter, and Poibeau}]{hamalainen2020automatic}
Mika H{\"a}m{\"a}l{\"a}inen, Niko Partanen, Khalid Alnajjar, Jack Rueter, and
  Thierry Poibeau. 2020.
\newblock Automatic dialect adaptation in finnish and its effect on perceived
  creativity.
\newblock In \emph{11th International Conference on Computational Creativity
  (ICCC’20)}. Association for Computational Creativity.

\bibitem[{Hermans(1985)}]{hermans2014manipulation}
Theo Hermans. 1985.
\newblock \emph{The manipulation of literature (routledge revivals): Studies in
  Literary Translation}.
\newblock Routledge.

\bibitem[{Hevner et~al.(2004)Hevner, March, Park, and Ram}]{hevner2004design}
Alan~R Hevner, Salvatore~T March, Jinsoo Park, and Sudha Ram. 2004.
\newblock Design science in information systems research.
\newblock \emph{MIS quarterly}, pages 75--105.

\bibitem[{Howcroft et~al.(2020)Howcroft, Belz, Clinciu, Gkatzia, Hasan,
  Mahamood, Mille, van Miltenburg, Santhanam, and
  Rieser}]{howcroft-etal-2020-twenty}
David~M. Howcroft, Anya Belz, Miruna-Adriana Clinciu, Dimitra Gkatzia, Sadid~A.
  Hasan, Saad Mahamood, Simon Mille, Emiel van Miltenburg, Sashank Santhanam,
  and Verena Rieser. 2020.
\newblock \href {https://www.aclweb.org/anthology/2020.inlg-1.23} {Twenty years
  of confusion in human evaluation: {NLG} needs evaluation sheets and
  standardised definitions}.
\newblock In \emph{Proceedings of the 13th International Conference on Natural
  Language Generation}, pages 169--182, Dublin, Ireland. Association for
  Computational Linguistics.

\bibitem[{Jordanous(2012)}]{Jordanous2012}
Anna Jordanous. 2012.
\newblock \href {https://doi.org/10.1007/s12559-012-9156-1} {A standardised
  procedure for evaluating creative systems: Computational creativity
  evaluation based on what it is to be creative}.
\newblock \emph{Cognitive Computation}, 4(3):246--279.

\bibitem[{Lederer(2003)}]{lederer2014translation}
Marianne Lederer. 2003.
\newblock \emph{Translation: The interpretive model}.
\newblock Routledge.

\bibitem[{van~der Lee et~al.(2019)van~der Lee, Gatt, van Miltenburg, Wubben,
  and Krahmer}]{van-der-lee-etal-2019-best}
Chris van~der Lee, Albert Gatt, Emiel van Miltenburg, Sander Wubben, and Emiel
  Krahmer. 2019.
\newblock \href {https://doi.org/10.18653/v1/W19-8643} {Best practices for the
  human evaluation of automatically generated text}.
\newblock In \emph{Proceedings of the 12th International Conference on Natural
  Language Generation}, pages 355--368, Tokyo, Japan. Association for
  Computational Linguistics.

\bibitem[{Mieskes et~al.(2019)Mieskes, Fort, N{\'e}v{\'e}ol, Grouin, and
  Cohen}]{mieskes2019nlp}
Margot Mieskes, Kar{\"e}n Fort, Aur{\'e}lie N{\'e}v{\'e}ol, Cyril Grouin, and
  Kevin~B Cohen. 2019.
\newblock Nlp community perspectives on replicability.
\newblock In \emph{Recent Advances in Natural Language Processing}.

\bibitem[{Mohankumar et~al.(2020)Mohankumar, Nema, Narasimhan, Khapra,
  Srinivasan, and Ravindran}]{mohankumar-etal-2020-towards}
Akash~Kumar Mohankumar, Preksha Nema, Sharan Narasimhan, Mitesh~M. Khapra,
  Balaji~Vasan Srinivasan, and Balaraman Ravindran. 2020.
\newblock \href {https://doi.org/10.18653/v1/2020.acl-main.387} {Towards
  transparent and explainable attention models}.
\newblock In \emph{Proceedings of the 58th Annual Meeting of the Association
  for Computational Linguistics}, pages 4206--4216, Online. Association for
  Computational Linguistics.

\bibitem[{Novikova et~al.(2017)Novikova, Dusek, Curry, and
  Rieser}]{novikova2017we}
Jekaterina Novikova, Ondrej Dusek, Amanda~Cercas Curry, and Verena Rieser.
  2017.
\newblock Why we need new evaluation metrics for nlg.
\newblock In \emph{2017 Conference on Empirical Methods in Natural Language
  Processing}, pages 2231--2242. Association for Computational Linguistics.

\bibitem[{Papineni et~al.(2002)Papineni, Roukos, Ward, and
  Zhu}]{papineni2002bleu}
Kishore Papineni, Salim Roukos, Todd Ward, and Wei-Jing Zhu. 2002.
\newblock Bleu: a method for automatic evaluation of machine translation.
\newblock In \emph{Proceedings of the 40th annual meeting of the Association
  for Computational Linguistics}, pages 311--318.

\bibitem[{Radford et~al.(2019)Radford, Wu, Child, Luan, Amodei, and
  Sutskever}]{radford2019language}
Alec Radford, Jeffrey Wu, Rewon Child, David Luan, Dario Amodei, and Ilya
  Sutskever. 2019.
\newblock Language models are unsupervised multitask learners.
\newblock \emph{OpenAI blog}, 1(8):9.

\bibitem[{Reiss(1989)}]{reiss1989text}
Katharina Reiss. 1989.
\newblock Text types, translation types and translation assessment.
\newblock \emph{Readings in translation theory}, 19771989.

\bibitem[{Reiter(2018)}]{reiter2018structured}
Ehud Reiter. 2018.
\newblock A structured review of the validity of {BLEU}.
\newblock \emph{Computational Linguistics}, 44(3):393--401.

\bibitem[{Roest et~al.(2020)Roest, Edman, Minnema, Kelly, Spenader, and
  Toral}]{roest-etal-2020-machine}
Christian Roest, Lukas Edman, Gosse Minnema, Kevin Kelly, Jennifer Spenader,
  and Antonio Toral. 2020.
\newblock \href {https://www.aclweb.org/anthology/2020.wmt-1.29} {Machine
  translation for {E}nglish{--}{I}nuktitut with segmentation, data acquisition
  and pre-training}.
\newblock In \emph{Proceedings of the Fifth Conference on Machine Translation},
  pages 274--281, Online. Association for Computational Linguistics.

\bibitem[{Talman et~al.(2019)Talman, Sulubacak, V{\'a}zquez, Scherrer,
  Virpioja, Raganato, Hurskainen, and Tiedemann}]{talman-etal-2019-university}
Aarne Talman, Umut Sulubacak, Ra{\'u}l V{\'a}zquez, Yves Scherrer, Sami
  Virpioja, Alessandro Raganato, Arvi Hurskainen, and J{\"o}rg Tiedemann. 2019.
\newblock \href {https://doi.org/10.18653/v1/W19-5347} {The university of
  {H}elsinki submissions to the {WMT}19 news translation task}.
\newblock In \emph{Proceedings of the Fourth Conference on Machine Translation
  (Volume 2: Shared Task Papers, Day 1)}, pages 412--423, Florence, Italy.
  Association for Computational Linguistics.

\bibitem[{Veale(2016)}]{Veale+2016+73+92}
Tony Veale. 2016.
\newblock \href {https://doi.org/doi:10.1515/9781501503993-004} {\emph{3. The
  shape of tweets to come: Automating language play in social networks}}, pages
  73--92. De Gruyter Mouton.

\bibitem[{Veale and Alnajjar(2015)}]{90fcde3a2b7e4813b3f74d3fb9b6b371}
Tony Veale and Khalid Alnajjar. 2015.
\newblock Unweaving the lexical rainbow: Grounding linguistic creativity in
  perceptual semantics.
\newblock In \emph{Proceedings of the Sixth International Conference on
  Computational Creativity}, pages 63 -- 70, United States. Brigham Young
  University.

\end{thebibliography}
\bibliographystyle{acl_natbib}

\end{document}